%% file: main.tex
\documentclass[10pt]{article} %

\usepackage{amssymb}            %
\usepackage{mathtools}          %
\usepackage{mathrsfs}           %
\usepackage{graphicx}           %
\usepackage{subcaption}         %
\usepackage[space]{grffile}     %
\usepackage{url}                %
\usepackage{lipsum}             %
\usepackage{multirow}
\usepackage{makecell}
\usepackage{array}
\usepackage{wrapfig}
\usepackage{blindtext}
\usepackage{booktabs} 

\usepackage{hyperref}
\usepackage{natbib}
\usepackage{authblk}
\usepackage{fullpage}

\usepackage{algorithm}
\usepackage[noend]{algpseudocode}
\algrenewcommand\algorithmicrequire{\textbf{Initialize}}
\algrenewcommand\algorithmicensure{\textbf{Input}}
\usepackage{enumitem}
\setlist{nolistsep}
\usepackage[dvipsnames,svgnames,table]{xcolor}

\title{Hierarchical Multi-agent Reinforcement Learning for Cyber Network Defense}

\author{Aditya Vikram Singh\textsuperscript{1}, Ethan Rathbun\textsuperscript{1}, Emma Graham\textsuperscript{2}, Lisa Oakley\textsuperscript{1}, \\ Simona Boboila\textsuperscript{1}, Peter Chin\textsuperscript{2}, Alina Oprea\textsuperscript{1}}

\affil{
$^{1}$Northeastern University\\
$^{2}$Dartmouth College
}

\date{}

\begin{document}

\maketitle 

\begin{abstract}
Recent advances in multi-agent reinforcement learning (MARL) have created opportunities to solve complex real-world tasks. Cybersecurity is a notable application area, where defending networks against sophisticated adversaries remains a challenging task typically performed by teams of security operators. In this work, we explore novel MARL strategies for building autonomous cyber network defenses that address challenges such as large policy spaces, partial observability, and stealthy, deceptive adversarial strategies. To facilitate efficient and generalized learning, we propose a hierarchical Proximal Policy Optimization (PPO) architecture that decomposes the cyber defense task into specific sub-tasks like network investigation and host recovery. Our approach involves training sub-policies for each sub-task using PPO enhanced with cybersecurity domain expertise. These sub-policies are then leveraged by a master defense policy that coordinates their selection to solve complex network defense tasks. Furthermore, the sub-policies can be fine-tuned and transferred with minimal cost to defend against shifts in adversarial behavior or changes in network settings. We conduct extensive experiments using CybORG Cage 4, the state-of-the-art MARL environment for cyber defense. Comparisons with multiple baselines across different adversaries show that our hierarchical learning approach achieves top performance in terms of convergence speed, episodic return, and several interpretable metrics relevant to cybersecurity, including the fraction of clean machines on the network, precision, and false positives.
\end{abstract}

%%%%%%%%%%%%%%%%%%%%%%%%%%%%%%%%%%%%%%%%%%%%%%%%%%%%%%%%%%%%%%%%
\input{introduction}

\input{related-work}
\input{problem-statement}
\input{method}

\input{experiments}

\input{conclusions}
%%%%%%%%%%%%%%%%%%%%%%%%%%%%%%%%%%%%%%%%%%%%%%%%%%%%%%%%%%%%%%%%

\subsubsection*{Acknowledgments}
\label{sec:ack}
This research was funded by the Defense Advanced Research Projects Agency (DARPA), under contract W912CG23C0031.

%%%%%%%%%%%%%%%%%%%%%%%%%%%%%%%%%%%%%%%%%%%%%%%%%%%%%%%%%%%%%%%%
%% NOTE: THIS MARKS THE END OF THE "MAIN TEXT"
%%%%%%%%%%%%%%%%%%%%%%%%%%%%%%%%%%%%%%%%%%%%%%%%%%%%%%%%%%%%%%%%

%%%%%%%%%%%%%%%%%%%%%%%%%%%%%%%%%%%%%%%%%%%%%%%%%%%%%%%%%%%%%%%%
%% Bibliography
%%%%%%%%%%%%%%%%%%%%%%%%%%%%%%%%%%%%%%%%%%%%%%%%%%%%%%%%%%%%%%%%
\bibliography{marl_refs}
\bibliographystyle{unsrt}

%%%%%%%%%%%%%%%%%%%%%%%%%%%%%%%%%%%%%%%%%%%%%%%%%%%%%%%%%%%%%%%%
%% Supplemental materials
%%%%%%%%%%%%%%%%%%%%%%%%%%%%%%%%%%%%%%%%%%%%%%%%%%%%%%%%%%%%%%%%

\newcommand{\beginsupplement}{%
        \clearpage
        \begin{center}
        \textbf{\LARGE Supplementary Materials}\\
        \vspace{0.1cm}
        \end{center}
        \setcounter{section}{0}
        \renewcommand{\thesection}{S-\arabic{section}}
        \setcounter{table}{0}
        \renewcommand{\thetable}{S-\arabic{table}}%
        \setcounter{figure}{0}
        \renewcommand{\thefigure}{S-\arabic{figure}}%
     }

\beginsupplement

\input{supplemental/game}

\input{supplemental/communication}
\input{supplemental/hmarl_diagram}

\input{supplemental/trafficControl}

\input{supplemental/transfer}

\end{document}

%% file: introduction.tex
\section{Introduction}

Cyber defense is critical in both private and public network infrastructures, which are frequently targeted by increasingly sophisticated external attackers with malicious intentions. In 2024, the number of security breaches has surpassed 10,000 and attackers constantly adapt their tools and strategies to evade existing defenses~\citep{Verizon}. The implications for national security are significant, as security breaches often lead to theft of intellectual property, compromise of sensitive information, and disruption of critical infrastructures.  Currently,  organizations employ teams of security professionals who constantly oversee the security of their networks and design the overall  security strategy using their domain expertise. While a range of machine learning (ML)  tools are available for detecting specific classes of attacks~\citep{180232,nelms2013execscent,antonakakis11,EXPOSURE,yen2013beehive,portfiler},  the advancement of deep reinforcement learning (DRL) presents an opportunity to automate the cyber defense strategy and reduce the burden on security operators.

Towards this goal, the technical cooperation program (TTCP), a collaborative working group including UK, USA, Canada, Australia and New Zealand, developed a series of CAGE challenges for advancing cyber defense~\citep{cage_challenge_announcement}. These challenges leverage the Cyber Operations Research Gym (CybORG), a simulated environment that can be used for creating realistic interactions (or games) between attackers and defenders on realistic network topologies.  These environments task defenders (blue agents) with monitoring and restoring compromised machines on a simulated enterprise network, to prevent external adversaries (red agents) from accessing critical assets. The first CAGE challenges  model a cyber game between a single blue agent and a single red agent, leading to the development of DRL techniques for training a blue agent interacting against a red agent~\citep{vyas2023automated, 10639381, hammar2024optimal}. The most recent CAGE 4 challenge~\citep{cage_challenge_4_announcement} models a team of multiple blue agents defending a distributed network, playing against multiple red agents compromising the network.  
Existing techniques for single defensive agents are either computationally expensive (by training a different agent for each red agent~\citep{vyas2023automated, 10639381}), do not generalize to new attackers, or require extensive, causal pre-processing which is intractable as the network size scales~\citep{hammar2024optimal}. Thus, they cannot be immediately applied to the multi-agent CAGE 4 environment, which requires new methods. Additional challenges for training multi-agent defenders in this environment include large policy spaces, partial observability of the network,
shared rewards among all blue agents, and playing against stealthy, deceptive adversarial strategies.

In this paper, we propose the first scalable multi-agent reinforcement learning (MARL) technique for automating defense in cyber security environments such as CAGE 4.  We formulate the problem as a decentralized, partially-observable Markov decision process (Dec-POMDP)~\citep{Oliehoek2016ACI} and propose two hierarchical strategies, H-MARL Expert and H-MARL Meta, each with their own advantages. Both methods decompose the complex cyber defense task into smaller sub-tasks, and 
train sub-policies for each sub-task using PPO enhanced with domain expertise. The difference between the methods is in the design of the master policy that coordinates the selection of the sub-policies at each time step. 
H-MARL Expert utilizes security domain expertise to define a top master policy based on well-established practices for cyber defense, and performs best in most of our experiments. However, there are situations when it is difficult to define a deterministic Expert policy. To address this issue, we propose  H-MARL Meta, that trains the master policy, and, thus, has the advantage of generalizing to new, unseen adversarial behavior.
Another insight in our design is that extended observation spaces including security indicators, such as presence of malicious files and malicious processes on a host, are beneficial in increasing the blue agent's ability to defend the network. We evaluate our methods against multiple baselines at different stages of the design process to motivate our methodology and design decisions. We further propose multiple relevant and interpretable metrics for cyber  defense, including ratio of uninfected hosts, false/true positive rates on host recovery, and number of adversarial impacts on hosts. Across these metrics our proposed hierarchical techniques display significant improvements over traditional MARL approaches.

To summarize, our contributions are: (1) scalable hierarchical multi-agent reinforcement learning methods for cyber defense; (2) a design guided by domain expertise to enhance the agents' observation space and decompose the complex cyber defense task into multiple sub-tasks; (3) evaluation in CybORG CAGE 4, a realistic cyber environment with partial observability and deceptive, stealthy adversaries; (4) empirical transferability of trained sub-policies after fine-tuning to new adversarial agents, and (5) multiple interpretable metrics for providing insights to security operators. The source code is available on GitHub~\footnote{https://github.com/adityavs14/Hierarchical-MARL}.

%% file: related-work.tex
\section{Related Work and Background}
Traditional cyber defenses, such as anti-virus  and network intrusion detection tools, leverage specific detection rules for thwarting existing attacks, but they are relatively easy to evade. To address their limitations, organizations employ security operators who perform ``threat hunting'' to detect novel attacks on their networks. A variety of machine learning (ML) tools are available for threat detection~\citep{antonakakis11,180232,EXPOSURE,nelms2013execscent,yen2013beehive,portfiler}, but the overall defensive strategy in most organizations is still manually designed. The advancement of DRL and MARL provides an opportunity to automate cyber defense strategies and improve the security of cyber infrastructures. 

The CAGE-4 challenge~\citep{cage_challenge_4_announcement} is a recent security environment aimed at encouraging research in autonomous cyber defense. It provides a cyber simulation of attacker and defender actions in realistic network topologies. CAGE-4 is a partially observable environment with multiple, decentralized blue agents defending the network by playing  against a  team of red agents performing various attacks over time. It can be modeled as a decentralized, partially observable Markov decision process (Dec-POMDP).  Dec-POMDPs~\citep{Oliehoek2016ACI} are a special class of MDP where multiple, independent and decentralized agents with incomplete observations interact to optimize a shared reward signal. Formally, a Dec-POMDPs is defined as the tuple $\mathcal{M} = (\mathcal{I}, \mathcal{S}, \mathcal{A}, \mathcal{T}, \Omega, \mathcal{O}, \mathcal{R}, b_o)$, where $\mathcal{I} = \{1 \cdots n\}$ is the set of $n$ agents, $\mathcal{S}$ is a finite set of states, $\mathcal{A} = \times_{i \in \mathcal{I}}A_i$ is the set of joint actions composed of individual actions $A_i$ for each agent $i$, $\mathcal{T}$ is the state transition function, $\Omega = \times_{i \in \mathcal{I}}\Omega_i$ is the set of joint observations, $\mathcal{O}: \mathcal{S} \times \mathcal{A} \rightarrow \Omega$ is the joint observation function, and $b_0$ is a distribution over initial states. 

Several methods have been developed for solving general Dec-POMDPs, including multi-agent PPO (MAPPO)~\citep{mappo,schulman2017proximal}, Q-MIX~\citep{qmix}, independent PPO (IPPO)~\citep{de2020independent}, or decision trees. However, these methods are often uninterpretable and struggle to converge in settings with large joint action spaces. These approaches have usually been applied to simpler 2-player environments~\citep{vyas2023automated,10639381,hammar2024optimal,wilson2024marlcyber}.
Prior works in the more complex CAGE-4 environment explore defenses based on heuristics~\citep{Kiely2025} or on traditional PPO algorithms~\citep{wang2024marlcyber}. In contrast, we implement an observation-enhanced  hierarchical learning method that is more adaptive than previous heuristic approaches, and more scalable than single-policy PPO architectures.

Hierarchical and meta-learning methods in reinforcement learning have led to adaptations of hierarchical MARL for different domains, such as multi-robot teamwork tasks~\citep{9196684,Fosong24,chang2023hierarchical} and complex navigation~\citep{Frans2018}.
To the best of our knowledge, our work is the first to study the design and capabilities of hierarchical MARL approaches in the cyber defense domain.

%% file: problem-statement.tex
\section{Problem Statement}\label{sec:problem}
In cyber security, red teams act as attackers who attempt to exploit network vulnerabilities and carry out malicious activities aimed at compromising the system. Blue teams are tasked with defending against red team opponents to secure the networks, while maintaining network operations.
In this work, we focus on the CybORG CAGE 4 cybersecurity MARL framework~\citep{cage_challenge_4_announcement}, which is a realistic environment that models cyber defense. We discuss several aspects of the environment below. For additional details, please see Section 1 in the supplemental materials, and the CAGE 4 description~\citep{cage_challenge_4_announcement}.

\vspace{6pt}
\noindent \textbf{Network topology.} Cyber networks are often segmented into operational enterprise networks that encompass multiple security zones depending on the proximity to critical resources. This setup leads to a multi-agent competitive environment, where each defender agent is protecting its own security zone(s), with the overarching team goal of defending the entire network. The CAGE 4 network consists of seven security zones (subnets), assigned to five blue agents. To increase robustness of defenses, the number of hosts in each zone and their services are randomized, with each zone having between 4-16 servers and user machines (or hosts). An additional network (Contractor) is completely undefended, so that the red team always maintains a presence in the network.

\vspace{6pt}
\noindent \textbf{Threat Model.}
The two teams are represented by multi-agent systems: defender (the blue team) and attacker (the red team). Defender and attacker have competing goals, while the agents on each team collaborate to achieve their goals.
The attacker's goal is to maximize its reward by degrading services available to users, represented by green agents, and compromising the critical Operational Technology (OT) service. The defender's goals are two-fold: maintain the security of the cyber network by reducing the adversarial presence, and minimize the operational impact on users.
We face a strong adversary, whose capabilities include stealth, phishing, propagating through the network, and the ability to discover blue agents' deception.  Red agents  maintain persistent presence on the Contractor network, which is the starting point of the attack and cannot be defended with blue actions.  Red agents scan machines for vulnerable services to exploit and propagate through the network. CAGE 4 implements both an \emph{aggressive service discovery} action, which is faster (1 time step), but has a high chance (0.75) of raising alerts, and a \emph{stealthy service discovery} action, which is slower (3 time steps), but less likely (0.25) to raise alerts. In addition to \emph{moving laterally} through the network by exploiting remote services, red agents can also \emph{spawn} with a given probability when a green user opens a phishing email or accesses a compromised service.
Furthermore, red agents can use the \emph{discover deception} action to determine if the blue team has installed decoy services on a specific host, and avoid to infect that host to maintain stealth. CAGE 4 implements a default deceptive red agent, called \emph{FiniteStateRedAgent}, but we also create our own red agents with more aggressive service discovery and even stealthier presence on the network to measure the generality of our defense. 
\input{challenges_table}

\vspace{6pt}
\noindent \textbf{Defensive actions.}
The blue team monitors the network for suspicious events, and detects and responds to attacks through the following actions: analyze a host looking for malware information; start a decoy service on a host (blue team is alerted when a red agent attempts to compromise the decoy service); block traffic to and from a specified security zone (at the  expense of disrupting the work of green agents); allow traffic to and from a specified security zone; remove malicious processes from a host; restore the host to an earlier secure state (temporarily making its services unavailable).

\vspace{6pt}
\noindent \textbf{Rewards.}
The reward scheme models a general-sum game where blue agents incur penalties when green agents are impacted due to degraded services becoming inaccessible. In addition, blue agents are penalized when red agents impact the critical OT security service, or when they use a costly action like Restore machine. The specific reward values depend on the \emph{mission phase} and are specified on the challenge page~\cite{cage_challenge_4_announcement}. Three mission phases are carried out throughout each episode, to reflect the changing criticality of security zones on current operations. Note that the reward includes the penalties incurred inside the contractor subnet, which cannot be defended. This additional reward should not affect the training process. For a fair comparison, the contractor reward is present in all the methods studied in this paper, including the baselines.

\vspace{6pt}
\noindent \textbf{Challenges.}
A series of features, described in Table~\ref{tab:challenges}, make the CybORG environment particularly realistic and challenging for training multi-agent blue defenders. The environment provides partial observability of red presence, as blue agents need to run monitor and analyse actions to discover compromised hosts, and these actions incur false positives and false negatives. The policy space is large, including a set of actions for each host on the network, and the observation space is memoryless. In addition, actions have variable duration, and all blue agents share a common reward, even though each of them protects a different part of the network. 

%% file: challenges_table.tex
\begin{table}[t]
\centering
\caption{Challenges of a realistic RL model for cybersecurity, with concrete examples from the CybORG  environment.}
\small
{%\def\arraystretch{1.1}

\begin{tabular}{p{0.26\linewidth} | p{0.68\linewidth}} \hline
\rowcolor[HTML]{EFEFEF}\multicolumn{2}{c}{\textbf{Environment}}\\\hline
Partially-observable & Blue agents receive incomplete information from the environment. Lacking access to the true state, agents monitor and analyse hosts to discover compromised hosts. Monitoring is noisy, affected by false positives and false negatives (depending on detection rate).\\ \hline
Memoryless & Once reported, alerts are not maintained.\\ \hline
Duration of actions & Actions take between one time step and 5 time steps in the environment.\\ \hline
Large policy space & Blue agents defend between 1 and 3 subnets, of up to 16 hosts each. This maps to an 92-242 action space, and an 82-210 observation space.\\ \hline
Shared reward & Blue agents are rewarded collectively, as a team, for defending the network. However, they only receive local observations.\\ \hline
\rowcolor[HTML]{EFEFEF}\multicolumn{2}{c}{\textbf{Adversary}}\\\hline 
Stealth &Low chance (0.25) of raising alerts via the Stealthy Service Discovery action. Ability to withdraw from a security zone after impact.\\\hline
Deception &Can detect and avoid decoy services.\\\hline
Ease of spreading & Appears via phishing emails, in addition to moving laterally through the network.\\\hline
Strong foothold & Can not be removed from the Contractor subnet, which is undefended. The red team scores about 60 points on average in this subnet.\\\hline
\end{tabular}
}
\label{tab:challenges}
\end{table}

%% file: method.tex
\section{H-MARL Methodology}
For complex real-world tasks, action spaces can grow intractably large as the problem scales, making task learning difficult for many standard reinforcement learning (RL) approaches~\citep{dulac2015deep}. These issues are compounded by the introduction of high-dimensional, noisy state spaces. Many hierarchical RL methods aim to solve these problems by breaking large action spaces down into smaller sub-tasks. However, learning these partitions online remains a challenging problem~\citep{Hutsebaut-Buysse2022}. Therefore, in this section, we introduce our hierarchical framework showing how domain expertise can be leveraged to solve issues of intractability in both the state and action spaces.

\begin{figure}[th]
%\begin{wrapfigure}[13]{L}{0.6\textwidth}
  \centering
 \includegraphics[width=0.8\textwidth,trim={1cm 6cm 0 6cm},clip]{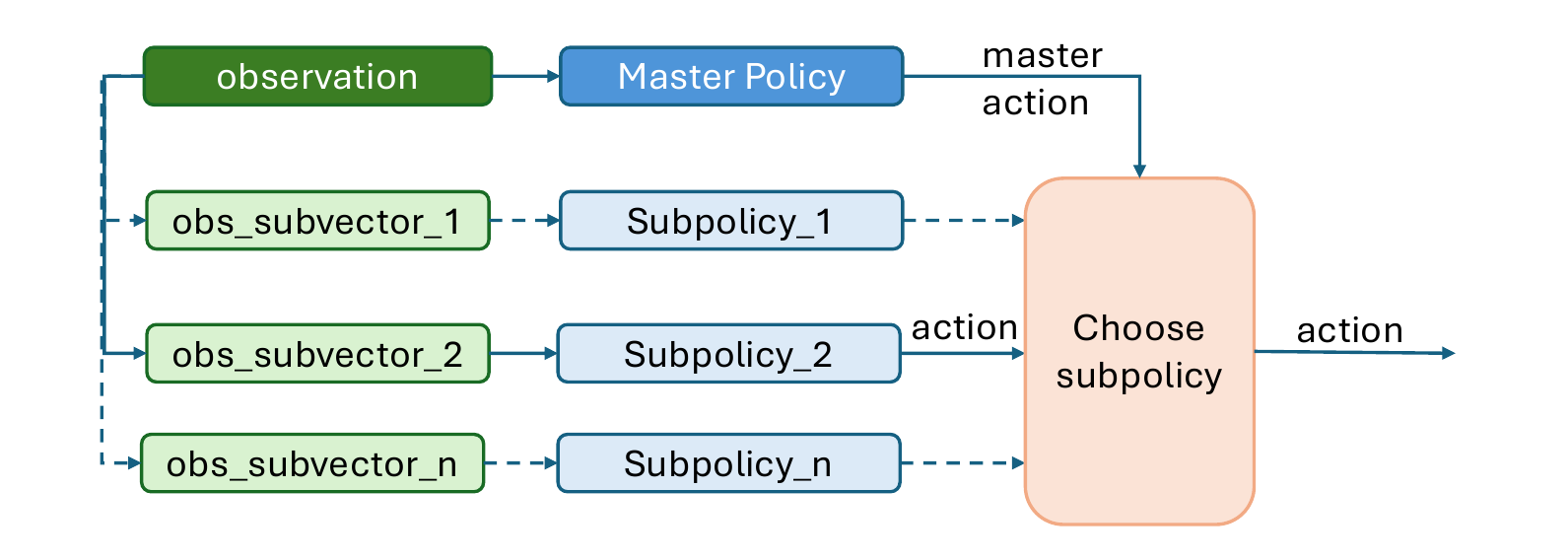}
     \caption{Hierarchical MARL. Upon receiving an observation, the master policy first chooses a sub-policy, which selects the final primitive action.}
     \label{fig:hrl_diagram_general}
%\end{wrapfigure}
\end{figure}

An overview of the hierarchical design is shown in Figure~\ref{fig:hrl_diagram_general}.
The action space $\mathcal{A}$ is split into $n$ smaller subsets, or classes, chosen using domain expertise. For example, the ``recover'' class refers to all primitive actions for removing processes and restoring machines on the network. Thus, each sub-policy handles one class of primitive actions that will be executed in the network. We define the action space of the master policy as a new set $\mathcal{A}_m$ comprised of meta-actions, where each meta-action corresponds to a different sub-policy. The master policy chooses a meta-action, and the associated sub-policy then samples a primitive action from its subset.  
In cyber domains, the agent must choose the machine to investigate or restore potentially from a list of thousands. In our design, these additional details, such as what machine to restore, will be abstracted away, under a single meta-action, significantly reducing the action space.
 
Under this formulation we first establish a master policy $\pi_m$ whose objective is to choose some meta-action $A_c \in \mathcal{A}_m$ given observation $o_t$ at time step $t$. We then assign each meta-action $A_c$ to a corresponding sub-policy $\psi_c$ whose goal is to choose the primitive action $a_t \in A_c$ given some input history $h_t$ at time step $t$. This primitive action $a_t$ is the final action executed in the environment. Under this design the master policy must learn the best policy $\pi_m : \mathcal{H} \rightarrow \mathcal{A}_m$ over meta-actions, while each sub-policy $\psi_c: \mathcal{H} \rightarrow A_c$ must learn the best policy over all actions in their respective meta-action class. Here $\mathcal{H}$ represents the set of possible observation histories. This reduces the larger, more complex task posed by the base Dec-POMDP into a much manageable set of sub-tasks.

We enhance each sub-policy's respective observations with transformation functions $f_c : \mathcal{O} \rightarrow \mathcal{O}_c$ for sub-policy observation spaces $\mathcal{O}_c$. In practice, these transformations are applied to observation histories. 
The transformation function reduces the observation space of each sub-policy by keeping only information relevant to their respective class of actions. For example, the sub-policy responsible for restoring machines only needs to know about the hosts that present clear indicators of compromise, rather than about all the alerts in the system.

\subsection{Hierarchical MARL Design} 
We now propose our methods H-MARL Expert and H-MARL Meta, which are designed to overcome a multitude of challenges induced by multi-agent training, such as environment and training instability. 
In particular, there are three key inter-dependencies that make learning in this setting difficult: (i) interdependence between each agent under a shared reward signal, (ii) between master and sub-policy performance, and (iii) between sub-policies under shared episodic returns. 

The first interdependence results in agents receiving rewards that are not related to actions they have taken. This is particularly challenging in our cyber defense setting as each agent interacts with disjoint sub-networks, but receives a shared reward considering the state of the whole network. This leads us to use IPPO~\citep{de2020independent} as the foundation of our approach, where each agent has a separate critic which only receives observations corresponding to their respective sub-network. Each agent is then trained in parallel along with its respective critic. This setup prevents the critics from being biased by occurrences outside their respective agent's sub-network, resulting in greater stability and less bias in each agent. 
Secondly, the performance of the master policy depends on the performance of each sub-policy --  poorly trained sub-policies can make otherwise optimal meta-actions sub-optimal -- resulting in a biased master policy. To overcome this, we utilize a two-phase training approach seen in Algorithm~\ref{alg:sub} and Algorithm~\ref{alg:master}. For both algorithms, the training is guided by the reward signal received from the environment (see Section~\ref{sec:problem}).

\vspace{6pt}
\noindent\textbf{Algorithm~\ref{alg:sub}: Sub-policy training (H-MARL Expert).} Our first method, H-MARL Expert, uses an expert master policy $\pi_E$ defined by domain expertise and only trains sub-policies for each agent, using Algorithm~\ref{alg:sub}. The H-MARL Expert Pipeline is presented in Figure~\ref{fig:pipeline-expert} from the Supplemental Materials.

At the start of each episode, we receive an initial observation from $\mathcal{M}$ and use it to initialize our history $h_t$. At each time step, $\pi_E$ then uses $h_t$ to choose the best meta-action $A_c$. 
Next, the sub-policy $\psi_c$ (corresponding to meta-action $A_c$) chooses a primitive action $a_t$ given its transformed history $f_c(h_t)$. This action is used to sample the next observation and reward from $\mathcal{M}$ which is stored in the replay memory $\mathcal{D}_c$ of $\psi_c$. 
Each policy is then updated with PPO on its respective replay memory $\mathcal{D}_c$. This design allows the sub-policies to address only their related tasks and train to near-optimal while avoiding instability caused by a trained master policy. Additionally, this allows us to solve our third form of interdependence, as the deterministic, static expert policy $\pi_E$ allows for stability in the training of each sub-policy.

\input{pseudocode}

\begin{figure}[t!]
     \centering
 \includegraphics[width=0.8\textwidth]{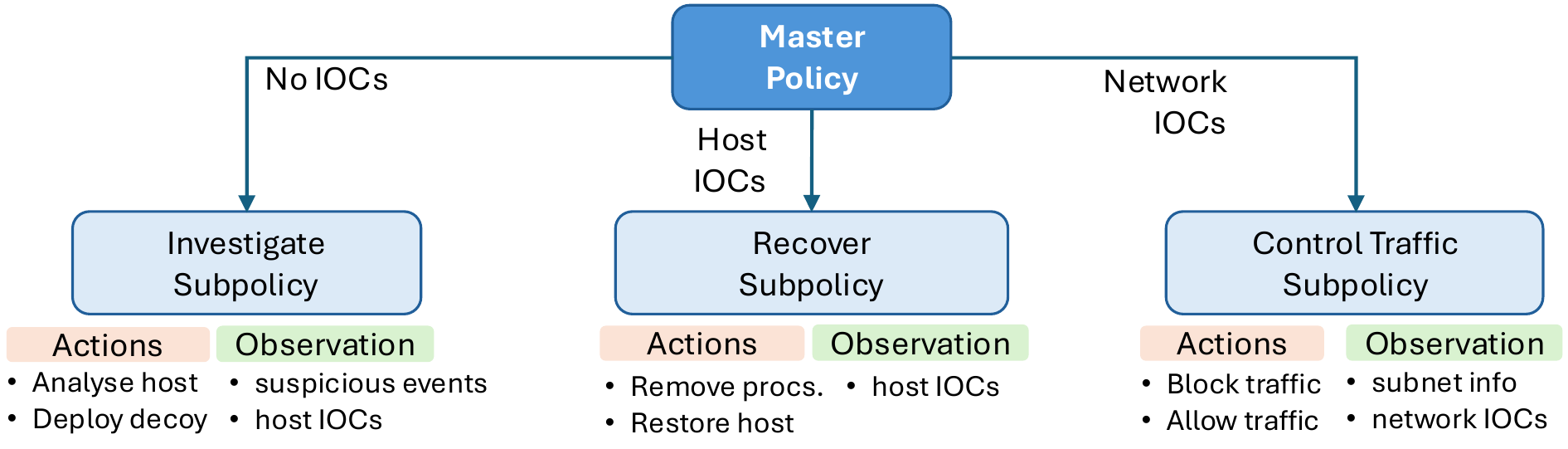}
     \caption{H-MARL in cybersecurity. The Expert Master Policy \emph{knows} this IOC-based partitioning, while the Meta Master Policy is \emph{learning} it using frozen subpolicies.}
     \label{fig:hrl_diagram}
\end{figure}

\vspace{6pt}
\noindent \textbf{H-MARL Expert in cybersecurity.} Figure~\ref{fig:hrl_diagram} illustrates the Expert master policy $\pi_E$ for partitioning the sub-tasks in  CybORG CAGE-4. We identify three types of sub-tasks: investigate host, recover host, and control traffic between zones. The state-level abstraction used to partition the tasks refers to the presence of indicators of compromise (IOCs) within an agent's security zone(s).
This partitioning is defined via Expert Rules, including: 
(1) If IOCs (malicious files) are detected on a host, the agent will choose the Recover subpolicy, which selects either to remove the malware or to restore the machine to a clean state;
(2) If network IOCs are detected, then the Control Traffic subpolicy is chosen;
(3) Otherwise, the agent will Investigate.

\vspace{6pt}
\noindent \textbf{Algorithm~\ref{alg:master}: Master policy training (H-MARL Meta).} Algorithm~\ref{alg:master} provides the second phase of training for the master policy. Here sub-policies $\{\psi\}_1^{k}$ trained with Algorithm~\ref{alg:sub} are kept frozen and just used to generate primitive actions. 
The H-MARL Meta Pipeline is presented in Figure~\ref{fig:pipeline-meta} from the Supplemental Materials.

We define a master policy $\pi_m$ called H-MARL Meta whose action space size is the number of sub-policies $k$. Similar to before, at each time step we sample an action $c_t \sim \pi_m(h_t)$ given updated history $h_t$. The sub-policy $\psi_{c_t}$ is  invoked to sample action $a_t$ to take a step in $\mathcal{M}$ given transformed history $f_c(h_t)$. The current history $h_t$, action $c_t$ and reward from $\mathcal{M}$ given action $a_t$ are then stored in the replay memory $\mathcal{D}_m$. Finally, the master policy $\pi_m$ is updated with PPO given trajectories stored in $\mathcal{D}_m$. 

\input{obs_space_design}

%% file: pseudocode.tex
\begin{figure}
%\vspace{-3em}
\begin{minipage}{0.48\textwidth}
\begin{algorithm}[H]
    \scriptsize
    \centering
    \caption{Sub-policy training (H-MARL Expert)}\label{alg:sub}
    \begin{algorithmic}[1]
    \Ensure Dec-POMDP $\mathcal{M}$, Expert policy $\pi_E$, Transformations $\{f_c\}_{c=1}^k$
    \Require Sub-Policies $\{\psi_c\}_{c=1}^{k}$, Replay Memories $\{\mathcal{D}_c\}_{c=1}^{k}$, Transformations $\{f_c\}_{c=1}^k$, Episode Length $T$, Iterations $N$
    \For{$i \gets 1,N$}
        \State Sample initial observation $o_0 \sim \mathcal{M}$
        \For{$t \gets 1,T$}
            \State Update history $h_t$ given observation $o_{t-1}$
            \State Sample $c \gets \pi_E(h_t)$ from expert policy 
            \State Sample $o_t, \; r_t \sim \mathcal{M}$ given action $a_t \sim \psi_c(f_c(h_t))$
            \State Store $(h_t, a_t, r_t)$ in $\mathcal{D}_c$ 
        \EndFor
        \For{$j \gets 1,k$}
            \State Update $\psi_j$ given $\mathcal{D}_j$ using PPO 
        \EndFor
    \EndFor
    \end{algorithmic}
\end{algorithm}
\end{minipage}
\hfill
\begin{minipage}{0.48\textwidth}
\begin{algorithm}[H]
    \scriptsize
    \centering
    \caption{Master policy training (H-MARL Meta)}\label{alg:master}
    \begin{algorithmic}[1]
    \Ensure  Dec-POMDP $\mathcal{M}$, Sub-Policies $\{\psi_c\}_{c=1}^{sp}$, Transformations $\{f_c\}_{c=1}^k$
    \Require Master Policy $\pi_m$, Replay Memory $\mathcal{D}_m$,  Episode Length $T$, Iterations $N$
    \For{$i \gets 1,N$}
        \State Sample initial observation $o_0 \sim \mathcal{M}$
        \For{$t \gets 1,T$} 
            \State Update history $h_t$ given observation $o_{t-1}$
            \State Sample $c_t \gets \pi_m(h_t)$ from master policy
            \State Sample $o_t, \; r_t \sim \mathcal{M}$ given action $a_t \sim \psi_{c_t}(f_c(h_t))$
            \State Store $(h_t, c_t, r_t)$ in $\mathcal{D}_m$ 
        \EndFor
        \State Update $\pi_m$ given $\mathcal{D}_m$ using PPO
    \EndFor
    \end{algorithmic}
\end{algorithm}
\end{minipage}
\end{figure}

%% file: obs_space_design.tex
\subsection{Observation Space Design} \label{sub:obs}

This section presents our enhancements to the observation space of blue agents and evaluate their ability to facilitate learning of a better defense strategy.
In cybersecurity environments, the state holds a lot of data, such as information on system configuration, processes, active sessions, etc. The observation is presented to the agent as a data structure with various fields, which may vary depending on the output of the action. This observation structure needs to be filtered and converted to a consistent vector representation to enable the use of deep reinforcement learning techniques. 

\begin{figure}[th]
    \centering
\includegraphics[width=0.7\linewidth]{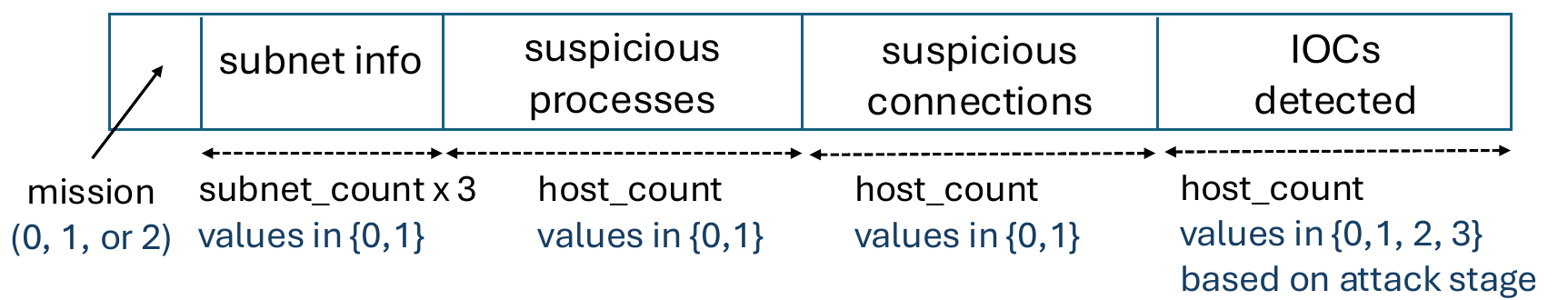}
    \caption{Observation space components. The basic CybORG observation is enhanced with IOCs.}
    \label{fig:obs_space_diagram}
\end{figure}

The basic observation vector of CybORG blue agents consists of the first four components from Figure~\ref{fig:obs_space_diagram}. The first bit represents the current mission phase. It is followed by a one-hot encoded vector with subnet-related information: what subnet(s) is the agent protecting, whether the traffic to/from other subnets is blocked, and which interfaces should be blocked based on the current mission phase restrictions. Next, the observation vector contains information on alerts detected with the Monitor action, using a 1/0 binary encoding to denote whether suspicious processes or connection events occurred on a host. 

\vspace{6pt}
\noindent \textbf{Enhancing observation with memory.}
Given that a single defense action can be taken per round, blue agents need a persistence mechanism to store alerts that have not been addressed yet. In CybORG, the Monitor action runs automatically at the end of each step but only reports \emph{new} events that have been raised on the current step. This requires us to maintain an updated observation history $h_t$ that keeps track of past events. Therefore, we add new events to the agent's history $h_t$ at each time step given a new observation $o_t$, and only remove these events from $h_t$ when they have been handled by a respective recover action.

\vspace{6pt}
\noindent \textbf{Enhancing observation with IOCs.}
Indicators of compromise (IOCs)~\citep{asiri2023} are signs or evidence of a cyber threat being present in the network. IOCs are generally classified in three categories: atomic (IP addresses, malware names, registry keys, process names, URLs, etc.), computed (hash of a malicious file), and behavioral~\citep{hutchins2011intelligence}. MITRE ATT\&CK (Adversarial Tactics, Techniques, and Common Knowledge)~\citep{strom2018mitre} is a comprehensive database of adversarial behaviors observed in real-world attacks.

One of our main insights for automating cyber defense is to extend the observation vector with IOC-related information per host, as illustrated in Figure~\ref{fig:obs_space_diagram}. We use two types of atomic IOCs: malicious file names that are placed on the victim machine and the IP address of the compromised host that issues service requests to a decoy service.
We also capture adversarial behaviors by prioritizing IOCs based on the attack phase.
A value of zero in the observation vector denotes that no IOC has been detected on the corresponding host. The other values differentiate between attack phases: priority 1 for IOCs detected during privilege escalation attempts, such as malicious files with root access; priority 2 for IOCs due to attacker's exploit actions (namely malicious files with user-level access); and priority 3 for IOCs due to the attacker's scanning activity (decoy accesses). 

\begin{figure}[th]
     \centering
\includegraphics[width=0.7\textwidth,trim={0 0.0cm 0 1.5cm},clip]{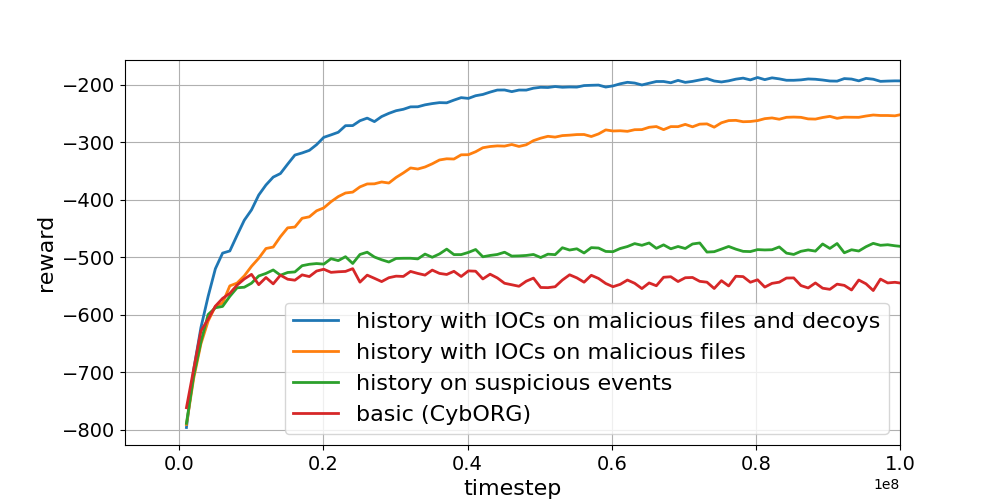}
     \caption{Blue team reward for different observation space designs. We incrementally add each of our enhancements to the observation space, to show their individual contribution. Incorporating history and IOCs provides a high performance boost. (MARL Decentralized training)}
     \label{fig:obs_design}
 \end{figure}

\vspace{6pt}
\noindent \textbf{Observation space evaluation.}
Figure~\ref{fig:obs_design} shows the contribution of each of our enhancements to the observation space design for a blue agent trained with a decentralized actor-critic PPO architecture, using the same hyper-parameters from Section~\ref{sec:experiments}. Compared to the basic CybORG observation, keeping track of history on suspicious events offers a small performance boost (12\% increase in reward). The biggest gain, however, comes from incorporating indicators of compromise related to malicious files (an additional increase in reward of 42\%). These files are detected when blue agents perform Analyse actions on hosts in their assigned subnet. Access to decoys, another clear indicator of adversarial behavior, further improves the defense strategy of the blue team by 11\%.  
Note that the H-MARL architecture can not be applied on the basic CybORG observation space. This is because task partitioning in H-MARL is based on indicators of compromise, which are not tracked in the original CybORG. In the rest of the paper, we consider the enhanced observation space with history, IOCs, and decoys for training all agents.

%% file: experiments.tex
\section{Experimental Evaluation}
\label{sec:experiments}
We now evaluate our proposed hierarchical MARL architecture in the CybORG CAGE 4 environment against different baselines, aiming to answer the following research questions: (i)  How effective is our H-MARL approach in protecting the network compared to other methods, against different adversaries (Sections~\ref{sec:results_performance} and~\ref{sec:experiments_adversaries})? (ii) Is it feasible to transfer previously trained sub-policies to learn new defense strategies (Section~\ref{sec:experiments_transfer})? (iii) Can we provide some interpretable insights to security operators related to the performance of our defenses (Section~\ref{sec:metrics})? 

\vspace{6pt}
\noindent \textbf{Training configuration.} Our experiments use the state-of-the-art actor-critic PPO algorithm~\citep{schulman2017proximal}. The actor and critic are represented by two feedforward neural networks with two hidden layers and 256 neurons per layer. The training hyper-parameters have been tuned to the following values: a learning rate of $5\times 10^{-5}$, a discount factor of 0.99, and a train buffer size of 1 million samples. The SGD algorithm uses the Generalized Advantage Estimation (GAE) function, a mini-batch size of 32,768 within each epoch, with 30 SGD iterations in each outer loop. Evaluation results are averaged across 100 randomized episodes, where each episode is 500 time steps long, and are accompanied by standard deviation information. The network topology in the CybORG environment is randomized and the models are trained on topologies with varying configurations, which ensures that they generalize to different network environments. The size of the network at initialization is randomly chosen between 32 and 128 hosts across 8 subnets, and up to 10 services are selected randomly to be placed on each host. All our models have been trained on 30 versions of the network, with separate workers collecting the experiences in their own network.

\vspace{6pt}
\noindent \textbf{Duration of actions.} For increased realism, actions take more than one step to execute. Some of the longer actions are Exploit Remote Services (4 steps) and Restore host (5 steps). In our implementation, while an action is in progress in the environment, we associate the ``in-progress'' state with a special observation and mask out other actions except sleep to guide PPO training.

\input{experiments/performance}

\input{experiments/transferability}

\input{experiments/expert_or_meta}

\input{experiments/interpretable_metrics}

%% file: experiments/performance.tex
\subsection{H-MARL Performance}
\label{sec:results_performance}

\textbf{MARL Baselines.} We compare the hierarchical architecture with two single-policy MARL paradigms~\citep{lyu2021contrastingcentralizeddecentralizedcritics}: Decentralized Training Decentralized Execution (DTDE), and Centralized Training Decentralized Execution (CTDE). 
The DTDE baseline uses an actor-critic architecture that learns a decentralized policy and critic for each of the agents locally (\emph{MARL Decentralized}).  
The CTDE baseline uses the centralized critic approach presented in MAPPO~\citep{mappo}.
Since each learning blue agent is required to guide its strategy based on the joint team reward, we augment the critic with state and actions of all blue agents on the team, while the actor only has access to local observations. Note that our \emph{MARL Centralized Critic} baseline uses the global state instead of incomplete agent observations to compute the joint value function. This is a reasonable assumption during training, 
in an effort to provide an unbiased and up-to-date critic for a strong baseline~\citep{lyu2021contrastingcentralizeddecentralizedcritics}. 

\vspace{6pt}
\noindent \textbf{H-MARL Methods.} 
We evaluate two hierarchical methods, H-MARL Expert and H-MARL Meta. \emph{H-MARL Expert} implements Algorithm 1, where the master policy is replaced with a rule informed by domain knowledge: \emph{If IOCs  are detected, Recover; otherwise, Investigate.} The goal is to recover right away to minimize the attacker's damage.
\emph{H-MARL Meta} implements Algorithms 1 and 2, following a curriculum style approach~\citep{Bengio2009}: the Recover and Investigate sub-policies pre-trained until convergence, are maintained fixed while training the master policy. 
\emph{H-MARL Collective} is an additional baseline that attempts to learn both the master and sub-policies from scratch, simultaneously. All the hierarchical variants use IPPO, in the decentralized actor-critic framework.

\begin{figure}[th]
 \centering
 \includegraphics[width=0.7\linewidth]{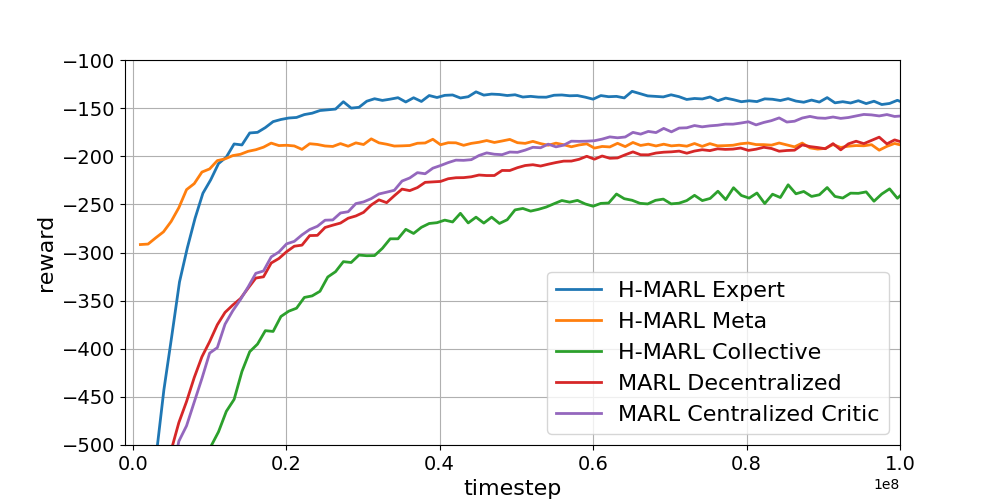}
    \caption{Average training return for all algorithms. H-MARL Expert is guided by a rule-based master policy and performs best. H-MARL Meta converges $3-5\times$ faster than MARL Decentralized.}
  \label{fig:hmarl_results}
 \end{figure}

The training process for the blue agents is presented in Figure~\ref{fig:hmarl_results}. 
The H-MARL Collective method performs the worst, in line with previous work~\citep{Frans2018} that  also observed the sub-optimal performance of updating both sub-policies and the master policy at the same time. Both single-policy methods MARL Decentralized and MARL Centralized Critic converge to a high reward, with a clear advantage for the shared critic method that estimates the return based on joint information.

As expected, the H-MARL Expert performs best ($-129.53$ reward), given that recovery actions are carried out promptly, before the attack amplifies. H-MARL Meta reaches a similar reward to MARL Decentralized ($-181.62$).  However, training a  master policy is significantly faster than training a single policy from scratch (about 3-5 times faster), as we are only tasked with choosing the correct sub-policy, rather than also learning the primitive sub-tasks, i.e., how to recover or investigate hosts.
H-MARL Meta invokes the fixed sub-policies that have been pre-trained with H-MARL Expert to take a step though the environment. The ability to \emph{learn} how to combine sub-tasks to solve higher-level tasks effectively with H-MARL Meta is particularly important in situations where defining an expert policy is difficult. We discuss such a scenario in Section~\ref{sec:expert_or_meta}.

\subsection{Evaluation against different adversaries}  
\label{sec:experiments_adversaries}
\input{experiments/reward_table}

We evaluated so far the performance of blue agents against the default red agent in CAGE 4, but we are interested in how the policy generalizes against other red attacks. We simulate red agent attackers that vary in their ability to circumvent the defense by employing more aggressive scanning, stealthier behavior, or a stronger focus on impacting critical services.
In the CybORG environment, red agents use finite state machine transitions to determine what actions to take for each known host. 
For a detailed description of the default finite state red agent and the transition matrix please see~\cite{cyborg_finite_state_red}. 

We consider four different finite state adversaries: (1) Default Red -- equal choice between the two available service discovery actions (stealthy and aggressive), and equal split between the two attack objectives (Impact and Degrade Service); (2) Aggressive Red -- always performs aggressive service discovery, a short duration action (1 time step) that has a high chance of being detected by blue agents (0.75); (3) Stealthy Red -- stealthy service discovery, characterized by low detection rate (0.25) and long duration (3 time steps);  (4) Impact Red -- fully committed to impacting the critical OT service, without attempting to degrade other services.

The evaluation results against these four different adversaries  are presented in Table~\ref{tab:algos}. Impact Red is the  strongest attacker collecting the highest reward against the blue team, in this zero-sum cyber game. Successful impact actions receive the highest reward of -10 during the second and third mission phases, more than any other network compromise. For comparison, users' failed access to degraded services only costs the blue team between -1 and -3. Aggressive Red is the second strongest opponent: the higher chance of being detected by blue agents is offset by the short duration of service discovery actions (1 time step), which enable this attacker to  carry out  more frequent attempts. Conversely, and somewhat non-intuitively, Stealthy Red poses less of a threat to blue agents due to the longer duration spent in covert scanning, before attempting to exploit.

H-MARL Expert is the best strategy against all four red agents and H-MARL Meta attains similar reward to the single policy baseline method (MARL Decentralized), but training converges significantly faster, as shown in the previous section. We confirm that H-MARL Meta performs better than  H-MARL Collective across all adversaries. Interestingly, it is also more effective than the MARL Centralized Critic method. 
Even though the centralized critic was trained jointly, using the global state, it is unable to utilize the shared information effectively on the individual actors during evaluation. This limitation of centralized critic policies has been previously investigated in the literature~\cite{lyu2021contrastingcentralizeddecentralizedcritics}.

%% file: experiments/reward_table.tex
\begin{table*}[t]
\centering
\caption{Mean evaluation reward against different adversaries, under different learning frameworks.}
\small
\resizebox{\textwidth}{!}{{\def\arraystretch{1.0}
\begin{tabular}{lcccc}
\hline
\multicolumn{1}{c}{\textbf{Opponent}} & \textbf{Default Red} & \textbf{Stealthy Red} & \textbf{Aggressive Red} & \textbf{Impact Red} \\ \hline
\textbf{MARL Decentralized} & $-179.8 \pm 92.98$ & $-165.8 \pm 53.45$ & $-227.93 \pm 87.98$ & $-247.15 \pm 71.53$ \\ 
\textbf{MARL Centralized Critic} & $-245.66 \pm 132.98$ & $-217.52 \pm 115.06$ & $-255.98 \pm 138.70$ & $ -332.96 \pm 108.13$ \\ 
\textbf{H-MARL Collective} & $-237.61 \pm 102.6$ & $-204.18 \pm 83.24$ & $-282.5 \pm 120.50$ & $-350.21 \pm 115.25$ \\ 
\textbf{H-MARL Expert} & $ \mathbf{ -129.53 \pm 44.60 }$ & $\mathbf{-99.54 \pm 38.28}$ & $\mathbf{-118.16 \pm 37.74}$ & $\mathbf{-173.17 \pm 64.04}$ \\ 
\textbf{H-MARL Meta} & $-181.62 \pm 65.85$ & $-184.76 \pm 91.40$ & $-207.66 \pm 86.48$ \& & $-278.78 \pm 97.76$ \\ 
\textbf{H-MARL Meta with fine-tuning} & $-186.24 \pm 81.28$ & $-162.51 \pm 57.97$ & n/a & $-264.96 \pm 81.53$ \\ \hline
\end{tabular}
}}
\label{tab:algos}
\end{table*}

%% file: experiments/transferability.tex
\subsection{H-MARL Transferability}
\label{sec:experiments_transfer}
In this section, we explore the possibility of knowledge transfer~\citep{Weinshall2018,pmlr-v80-wei18a,zhuang2020transferlearning,NekoeiBCC21} -- given a defense strategy that has already been trained, can we use it to accelerate the learning of a new defense that protects against a different adversary?
The adversaries in this study are stationary randomized agents, which consistently apply the same probabilistic rules to make decisions. We leave as future work the study of H-MARL transferability to evolving adversaries that adapt their policies based on opponent behavior or environment feedback. 
We adapt the pre-trained sub-policy models to new attacks via fine-tuning, a well-established powerful transfer method for deep models~\citep{zhuang2020transferlearning,Sun2019} that is  significantly less costly than training from scratch. In our current design, we only fine-tune the sub-policies for a few iterations. % (between 5 and 10). 
Next, we fully train a new master policy, using the tuned sub-policies. 

\begin{figure}[th]
    \centering
   \includegraphics[width=0.7\linewidth]{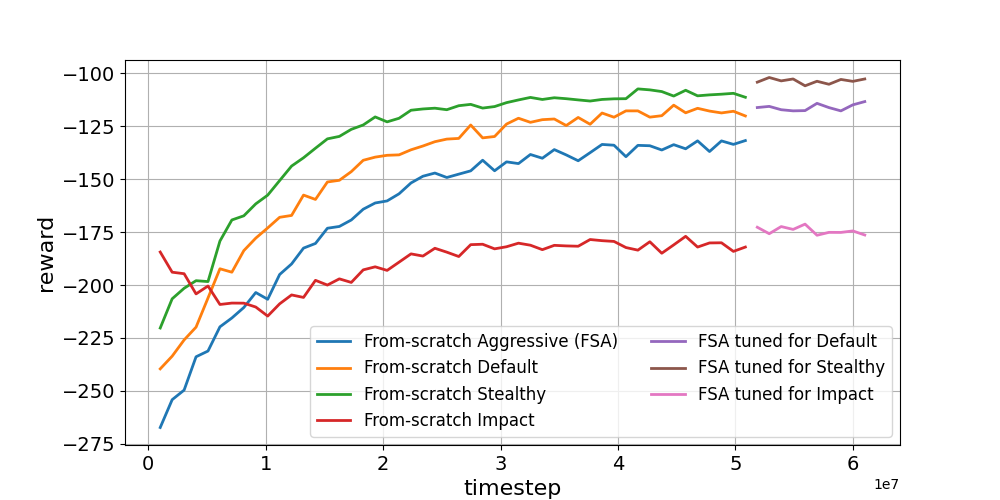}
    \caption{The Investigate sub-policy pre-trained against Aggressive Red is fine-tuned separately against other red agents, performing similarly to training from scratch. }\label{fig:transfer_investigate}
\end{figure}

Figure~\ref{fig:transfer_investigate} presents  the fine-tuning results of the Investigate sub-policy, pre-trained against Aggressive Red, the average-performing attacker. A short fine-tuning is enough to adapt to a new adversary, resulting in learning curves similar to those obtained when training from scratch. 
The fine-tuning results for the Recover sub-policy are included in Section~\ref{sec:supl_transfer} the Supplemental Material. We note that Investigate learns different strategies for each adversary (due to different patterns of suspicious events), while Recover is relatively agnostic to the attack (due to working on clear indicators of compromise) and can be directly reused against other red agents. 

Last row in Table~\ref{tab:algos} shows the evaluation results of the master policy that was trained with the fine-tuned sub-policy models. In this case, the master policy performs comparably to (and sometimes even better than) H-MARL Meta that uses sub-policies trained from scratch. For instance, against the Stealthy Red agent, the fine-tuned H-MARL Meta policy attains an average reward of $-162.51$, compared to an average $-184.76$ reward when training from scratch. Our empirical results are supported by previous research, which has shown that pre-trained deep learning models have been proven to generalize better than randomly initialized ones~\citep{Erhan2010,Sun2019}.

%% file: experiments/expert_or_meta.tex
\subsection{What method to use: Expert or Meta?}
\label{sec:expert_or_meta}
In our previous experiments, we have seen that H-MARL Expert performs best, as it is guided by a deterministic master policy generated from domain knowledge, specifically: \emph{If host IOCs are present, invoke the Recover sub-policy; otherwise Investigate.} To show that such expert rules do not generalize in all situations, we consider a scenario with a third sub-policy, Control Traffic, in which a defender may choose to block traffic between subnets. 

\begin{figure}[th]
     \centering
      \includegraphics[width=0.7\linewidth]{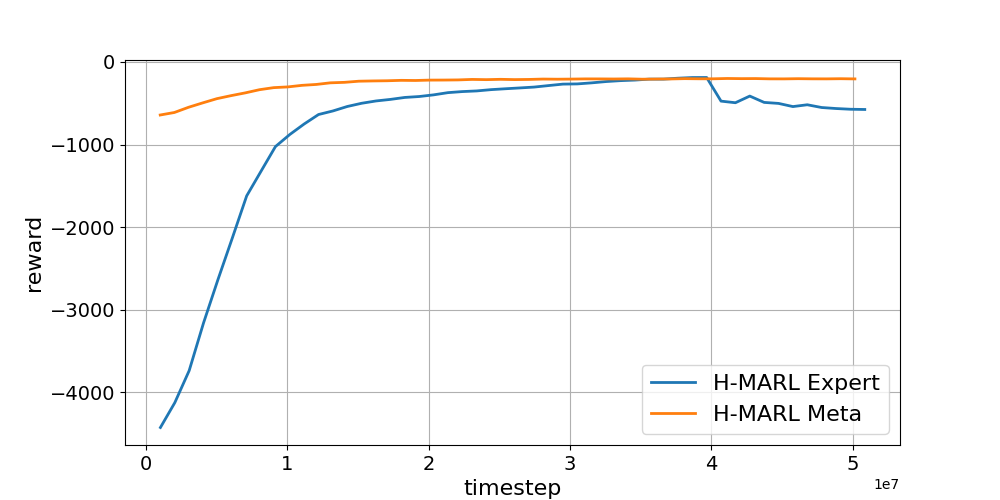}
     \caption{H-MARL with 3 sub-policies: Investigate, Recover, and Control Traffic. Due to the probabilistic expert rule, H-MARL Expert method is unstable, while H-MARL Meta performs well, converging fast to a stable performance.}
     \label{fig:expormeta}
\end{figure}

However, block actions incur penalties for preventing the remote activity of green users, rendering a deterministic blocking rule unfavorable. We instead experimented with a probabilistic master policy: \emph{If host IOCs are present, invoke the Recover sub-policy; otherwise split 75\%-25\% between Investigate and Control Traffic.} With this probabilistic rule, the H-MARL Expert method becomes unstable, while H-MARL Meta performs better and converges quickly to a stable performance (see Figure~\ref{fig:expormeta}). This demonstrates the importance of learning the master policy and is discussed in more detail in Section~\ref{sec:trafficcontrol} of the Supplemental Material.

%% file: experiments/interpretable_metrics.tex
\input{experiments/metrics_table}

\subsection{Interpretable Metrics}
\label{sec:metrics}

In cybersecurity applications, the defender has two main objectives: keep the network secure and maintain operational workflows. Total discounted rewards is a natural metric for training and measuring performance in reinforcement learning, however, a single-valued reward provides little interpretable information to security operators. To address this problem, we propose a set of interpretable metrics and analyze their relationship to the existing reward computation.
The new metrics provide insight on defense performance from three perspectives -- network security, effectiveness of recoveries, and impact on operations -- as follows: 

\begin{itemize}[itemsep=0mm]
    \item Network Security Posture:
    \begin{itemize}
        \item \textit{Clean Hosts}: Fraction of hosts with no red presence (from total hosts in the network).
        \item \textit{Non-Escalated Hosts}: Fraction of hosts with no red root sessions.
    \end{itemize}
    \item Recovery Metrics:
    \begin{itemize}
        \item \textit{Mean Time to Recover}: 
        Mean number of consecutive steps spent in a compromised state.
        \item \textit{Useful Recoveries}: Recoveries performed on infected machines (true positives, TP).
        \item \textit{Wasted Recoveries}: Recoveries performed on clean machines (false positives, FP).
        \item \textit{Recovery Error}: 
        $\text{Err = FP /  (TP + FP)}$.
        \item \textit{Recovery Precision}: 
        $\text{TP /  (TP + FP) = 1-Err}$.
    \end{itemize}
    \item Operational Metrics:
    \begin{itemize}
        \item \textit{Red Impact Count}: Number of times the OT service is impacted becoming unavailable. 
    \end{itemize}
\end{itemize}

We evaluate our blue agent strategies according to these metrics in Table \ref{tab:interpret}. We observe a number of insights that are not evident by comparing policies using solely the reward metric. For instance, MARL Decentralized performs more recoveries than other policies, but its recovery precision is only $0.27$. While H-MARL Meta and MARL Decentralized have similar rewards, H-MARL Meta has a much better recovery precision of $0.61$. 
Moreover, red agents are less successful at impacting the critical services with H-MARL Meta, compared to MARL Decentralized. 
These indicators show that H-MARL Meta is a more effective defense strategy than the single-policy approach (MARL Decentralized), despite having similar reward. H-MARL Expert has the highest recovery precision across all policies, as its expert master policy selects the Recovery sub-policy when IOCs are present on hosts, a strong indication of host compromise.  Our analysis demonstrates the need of using the reward signal in conjunction with other  metrics that are relevant in the cyber domain.
To better align the reward with the defender's goals, one can also incorporate these new metrics in the reward.

%% file: experiments/metrics_table.tex
\begin{table*}[ht]
\centering
\caption{Interpretable metrics for various blue strategies against Default Red, 100-episode averages.}
\label{tab:interpret}
\resizebox{\textwidth}{!}{%
\begin{tabular}{l|l|l|l|l|l|l|l|l|l}\hline
\textbf{\begin{tabular}[c]{@{}l@{}}Blue Strategy\end{tabular}}        
& \begin{tabular}[c]{@{}l@{}}\textbf{Clean} \\ \textbf{Hosts}\\ \textbf{(ratio)}\end{tabular} 
& \begin{tabular}[c]{@{}l@{}}\textbf{Non-Escalated} \\ \textbf{Hosts}\\\textbf{(ratio)}\end{tabular} 
& \textbf{\begin{tabular}[c]{@{}l@{}}Mean\\Time to\\Recover\end{tabular}} 
& \begin{tabular}[c]{@{}l@{}}\textbf{Useful}\\\textbf{Recoveries}\\\textbf{(true positives)}\end{tabular}
& \begin{tabular}[c]{@{}l@{}}\textbf{Wasted}\\\textbf{Recoveries}\\ \textbf{(false positives)}\end{tabular}
& \begin{tabular}[c]{@{}l@{}}\textbf{Recovery} \\ \textbf{Precision} \end{tabular} 
& \begin{tabular}[c]{@{}l@{}}\textbf{Recovery} \\ \textbf{Error}\end{tabular}
& \textbf{\begin{tabular}[c]{@{}l@{}}Red\\Impact \\ Count\end{tabular}} 
& \textbf{\begin{tabular}[c]{@{}l@{}}Reward\end{tabular}} \\ \hline
\textbf{\begin{tabular}[c]{@{}l@{}}H-MARL Expert\end{tabular}}  & 0.81 & 0.99 & 44.76 & 6.07 & 2.2 & 0.73 & 0.27 & 0.88 & {\bf-129.53} \\ \hline
\textbf{\begin{tabular}[c]{@{}l@{}}H-MARL Meta\end{tabular}} & 0.77 & 0.97 & 62.63 & 5.79 & 3.74 & 0.61 & 0.39 & 1.14 & -181.62 \\ \hline
\textbf{\begin{tabular}[c]{@{}l@{}}H-MARL Collective\end{tabular}} & 0.85 & 0.97 & 35.52 & 3.41 & 5.79 & 0.37 & 0.63 & 1.93 & -255.56 \\ \hline
\textbf{\begin{tabular}[c]{@{}l@{}}MARL Decentralized\end{tabular}} & 0.77  & 0.97 & 56.08 & 10.45 & 27.99 & 0.27 & 0.73 & 1.54 & -180\\ \hline                                                       
\end{tabular}%
}
\end{table*}

%% file: conclusions.tex
\section{Conclusion}

We propose novel hierarchical multi-agent reinforcement learning (MARL) strategies to train multiple blue agents tasked with protecting a network against red agents. Our H-MARL  strategy decomposes cyber defense into multiple sub-tasks, trains sub-policies for each sub-task guided by domain expertise, and finally trains a master policy to coordinate sub-policy selection. We evaluated our proposed hierarchical methods (Expert and Meta) and compared them against standard decentralized and centralized MARL in a realistic cyber security environment, CybORG CAGE 4.
%We demonstrate that our hierarchical policy converges faster than a single PPO policy and achieves higher reward than a hierarchical policy that trains the master policies and sub-policies collectively. 
We demonstrated that our hierarchical method converges faster than a single PPO policy and generalizes across various red agent behavior, while H-MARL Expert consistently performs better than the baselines. 
%We show the generality of our approach across various red agent behavior and introduce a number of interpretable metrics to provide insights to security operators. %We leave to future work the question of reward shaping to account for multiple defender objectives. 

% Emma - original conclusion is below 

%Defending cyber networks against advanced adversaries is a challenging task, currently performed by teams of security operators.
%We propose a novel hierarchical multi-agent RL strategy with the goal of  training a set of multiple blue agents tasked with protecting a network against a set of red agents. Our H-MARL Meta strategy decomposes cyber defense into multiple sub-tasks, trains sub-policies for each sub-task guided by domain expertise, and finally trains a master policy to coordinate sub-policy selection. We evaluate our proposed method and compare it against both centralized and other hierarchical methods in a realistic cyber security environment, CybORG CAGE 4.  We demonstrate that H-MARL Meta converges faster than a single PPO policy, and achieves higher reward than a hierarchical policy that trains the master policies and sub-policies collectively. We show the generality of our approach to various red agent behavior, and introduce a number of interpretable metrics for providing  insights to security operators. We leave to future work the question of reward shaping to account for multiple defender objectives. 

%% file: supplemental/game.tex
\section{The Cyber Game}
This section presents additional details on the cyber game simulated in CybORG CAGE 4~\footnote{https://github.com/cage-challenge/cage-challenge-4/tree/main}. 
For increased realism, users on enterprise machines are represented by green agents, which are present on every host. Green agents randomly access local and remote services, such as SSHD, MySQL, FTP, etc. Upon compromise, red agents are able to \emph{degrade} these services, preventing users from completing their work or slowing them down.
The red team can attack the network through several actions: scan the network to discover new hosts; scan a host to discover active services; exploit a vulnerability to compromise a host; escalate privileges on a compromised node to gain root access; degrade user experience of green agents; impact (stop) the critical Operational technology (OT) service; discover deception (i.e., probe a host to determine if it is running decoy services).

The action sets for green agents (users), blue agents (defenders) and red agents (attackers) are presented in Table~\ref{tab:actions}.
Both blue and red agent actions have a specified time duration, ranging from one time step (block/allow traffic) to five time steps for longer actions (restore host). Agents are not allowed to choose another action before the previous action had completed.

\input{supplemental/actions_table}

The red agent in CAGE 4 is rather sophisticated and randomized, not deterministic, and is modeled by:
\begin{itemize}
\item State transition diagram: The adversary follows a state transition diagram with eight states and nine transitions (that is, actions) that defines the allowed subset of actions based on the state of the host. 
\item State transition probability matrix: The adversary’s actions are randomized, using a state transition probability matrix to choose among the subset of possible actions in each state. For example, in State S (i.e., a Service was discovered on a host), the Red agent can choose to Discover Remote Systems with 0.25 probability, Discover Deception with 0.25 probability, or Exploit Remote Service with 0.5 probability.
\end{itemize}

We vary the probabilities in the state transition probability matrix to create additional red agent variants and cover a wider range of Red behaviors (fully Aggressive, Stealthy, or Impact). These new attack vectors  explore how two important characteristics of any cyber attack, namely the speed/stealth of discovering vulnerabilities, and the attacker's objective affect the attack success.

%% file: supplemental/actions_table.tex
\begin{table}[ht]
\centering
\caption{Action sets for green agents (users), blue agents (defenders), and red agents (attackers).}
\small
{%\def\arraystretch{1.1}

\begin{tabular}{p{0.26\linewidth} | p{0.65\linewidth}} \hline

\rowcolor[HTML]{EFEFEF}\multicolumn{2}{c}{\textbf{Green actions}}\\\hline
Green Access Service & Communicate with a server in the local zone or another zone. It has a small chance of being flagged as malicious by Monitor.\\\hline
Green Local Work &Do work on the local host without communicating. It has a small chance that this action results in red gaining a foothold on the host (e.g. as a result of a Phishing email). Also, it has a small chance of being flagged as malicious by Monitor.
\\\hline
\rowcolor[HTML]{EFEFEF}\multicolumn{2}{c}{\textbf{Blue actions}}\\\hline
Monitor & Collection of information about flagged malicious activity on the network.\\ \hline
Analyse & Collection of further information relating to malware files, on a specific host.\\ \hline
Deploy Decoy & Setup of a decoy service (type randomly assigned) on a specified host. Blue is alerted if Red accesses this service.\\ \hline
Remove & Attempts to stop all processes identified as malicious by the monitor action.\\ \hline
Restore & Restoring a system to a known good state.\\ \hline
Block Traffic & Block traffic to and from the specified zone.\\ \hline
Allow Traffic & Allow traffic to and from the specified zone.\\ \hline

\rowcolor[HTML]{EFEFEF}\multicolumn{2}{c}{\textbf{Red actions}}\\\hline 
Discover Remote Systems & Discovers new hosts/IP addresses in the network through active scanning using tools such as ping. \\ \hline
Aggressive Service Discovery & Discovers responsive services on a selected host by initiating a connection with that host.\\ \hline
Stealth Service Discovery & Same as Aggressive Service Discovery but slower and quieter (i.e., lower chance of raising alerts). \\\hline
Exploit Network Services & Attempts to exploit a specified service on a remote system.\\\hline
Privilege Escalate & This action escalates the agent’s privilege on the host.\\\hline
Impact & This action disrupts the performance of the network and fulfils red’s objective of denying the operational service. \\\hline
Degrade Services & If red has root privileges on a host, it may degrade the user experience for a green agent.  \\\hline
Discover Deception & Probe a host to determine if it is running decoy services.\\\hline
Withdraw & Remove red presence from target host.\\\hline

\end{tabular}
}
\label{tab:actions}
\end{table}

%% file: supplemental/communication.tex
\section{Communication and Cooperation}
\label{sec:experiments_communication}
In this section, we  explore other possible extensions that can help the defense strategy, focusing on communication among agents. In the CybORG CAGE 4 environment, blue agents are facing a challenging adversary, who can move through the network in two ways: (1) phishing emails, and (2) active scanning of hosts and services. Each red agent conducts scanning activity mostly within its own assigned subnet(s), and rarely reaches remotely into other subnets. This partitioning is useful from a scaling perspective, to limit the observation and action spaces. However, it also leads to limited compromise attempts that cross subnet boundaries. Thus, each defender can focus its efforts on its own assigned subnet(s), requiring little communication or coordination with other blue agents. Still, communication can be useful in other game settings, to send information about network-level indicators of compromise, such as malicious file names, the hash of a malicious file, or a compromised IP.
\begin{figure}[th]
    \centering
\includegraphics[width=0.7\linewidth]{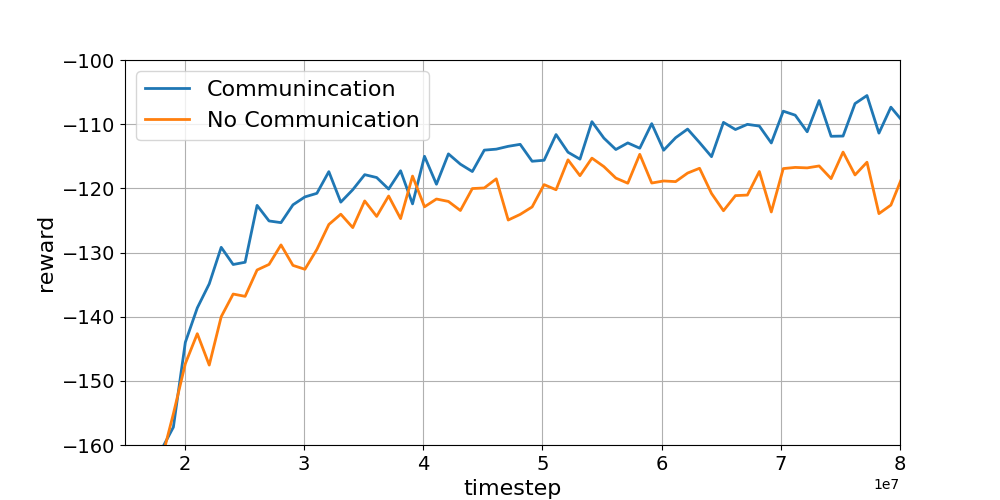}
    \caption{Blue agents use 8-bit messages to warn other team members of potential compromised hosts. This communication strategy shows some benefit over the case when no communication is used. (MARL Decentralized training)}
    \label{fig:communication}
\end{figure}

As a case study, we implemented a red agent that chooses external scanning in 50\% of the time (once it becomes aware of another subnet), and a blue agent that relies heavily on decoys (90\% of blue actions) to detect the adversary during the scanning phase of the attack.
Blue agents broadcast 8-bit messages encoding which remote host is accessing their decoys to warn other agents of potential attackers. We uniquely identify the compromised hosts with 3 bits for the subnet number (1-7), and 4 bits for the host index (0-15). Each blue agent decodes the message to check if it refers to hosts from its own subnet. If so, the message provides an indicator of compromise that will be added to the observation vector. Figure~\ref{fig:communication} shows that there is some benefit of using this method of communication. However, the benefit is small, due to other factors, such as phishing, stealth, and false positives of the Monitor actions. 

%% file: supplemental/hmarl_diagram.tex
\section{H-MARL Pipeline}
\label{sec:pipeline}
Meta actions describe a class of actions (e.g., Restore, Investigate, Control Traffic), which the master policy selects. This partitioning abstracts away additional details, such as what machine to restore, making training and generalization easier for the master policy. Sub-policies describe the policies that choose the primitive action ultimately executed in the environment (e.g., restore host 13). In effect the master policy chooses a meta action, and then samples a primitive action from the respective sub-policy.
\begin{figure}[th]%
    \centering
\includegraphics[width=1\textwidth]{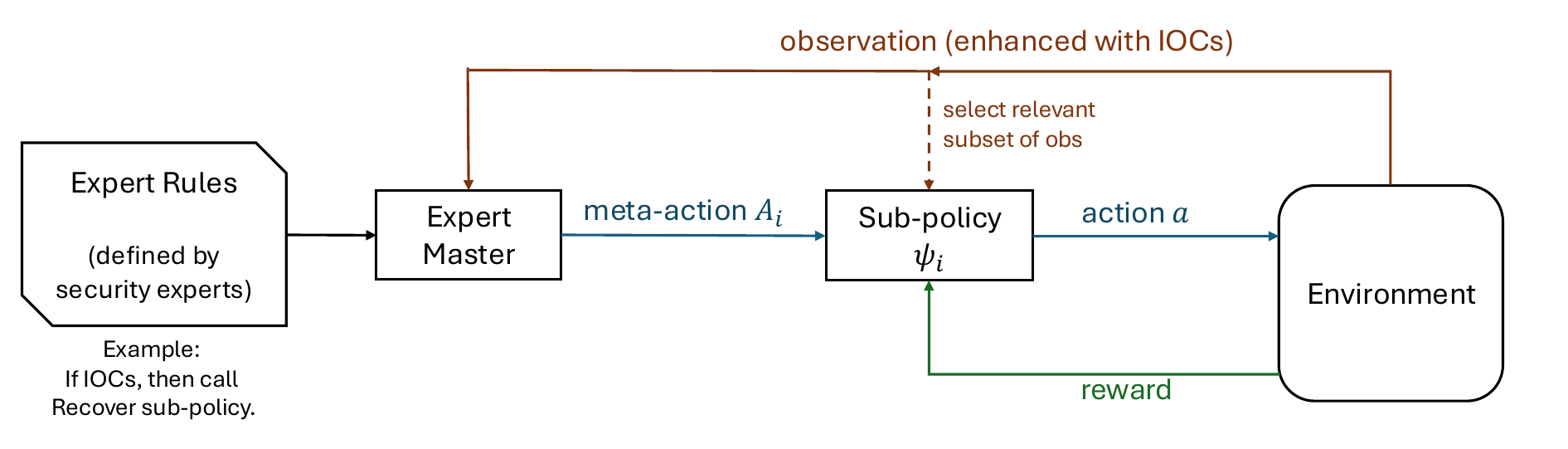}
   \caption{H-MARL Expert Pipeline. The Master uses expert rules to choose a sub-policy that steps through the environment. Sub-policy training is guided by the reward from the environment.}
   \label{fig:pipeline-expert}
\end{figure}

An overview of the \textbf{H-MARL Expert pipeline} is shown in Figure~\ref{fig:pipeline-expert}. Upon receiving an observation the expert master uses pre-defined rules to choose a meta-action indexed by $i$ The observation is then processed into the respective observation space of sub-policy $i$, which chooses the final primitive action to step through the environment. This approach aims to learn sub-policies that are specialized for a single task. 

\begin{figure}[th]%
    \centering
\includegraphics[width=0.85\textwidth]{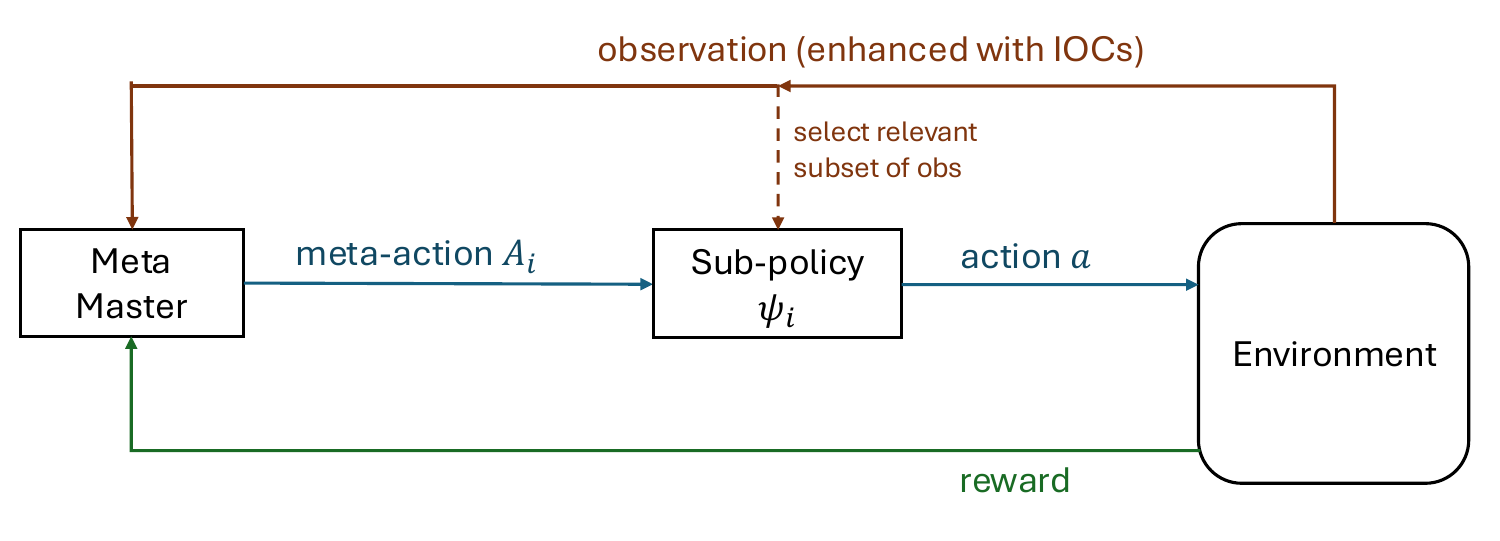}
   \caption{H-MARL Meta Pipeline. The Master learns a probability distribution over meta-actions. The Master training uses frozen sub-policies and is guided by the reward from the environment.}
   \label{fig:pipeline-meta}
\end{figure}

An overview of the \textbf{H-MARL Meta pipeline} is shown in Figure~\ref{fig:pipeline-meta}.
The meta master policy learns to reason on a higher-level about the decisions it can make by using state abstractions (e.g. are IOCs  present in the  network?). During the training of the meta master, the sub-policies are kept frozen. The master learns a probability distribution over the meta-actions, guided by the reward from the environment.

%% file: supplemental/trafficControl.tex
\section{Traffic Control} 
\label{sec:trafficcontrol}
Our next case study explores the use of Block and Allow Traffic actions to control the access between security zones. The H-MARL architecture consist of a Master policy and three sub-policies: Investigate, Recover, and Control Traffic. We extended the observation space with network-level indicators of compromise -- blue agents communicate whether their assigned subnet(s) contain any IOCs -- enabling each agent to have a global view on the network, and facilitating the training of the Control Traffic sub-policy.
\begin{figure}[th!]
    \centering
     \includegraphics[width=0.7\linewidth]{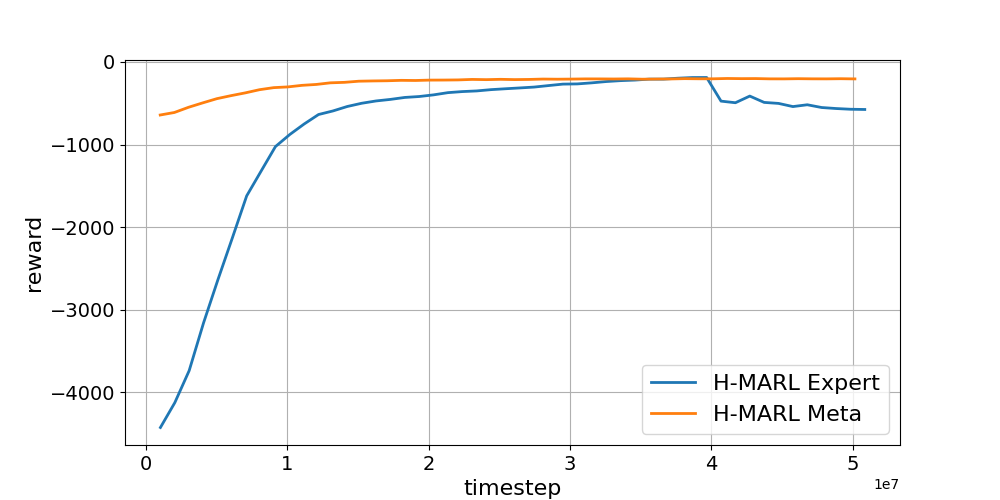}
    \caption{H-MARL with 3 sub-policies: Investigate, Recover, and Control Traffic. Due to the probabilistic expert rule, H-MARL Expert method is unstable, while H-MARL Meta performs well, converging fast to a stable performance.}
    \label{fig:traffic}
\end{figure}

\begin{figure}[th!]
    \centering
     \includegraphics[width=0.7\linewidth]{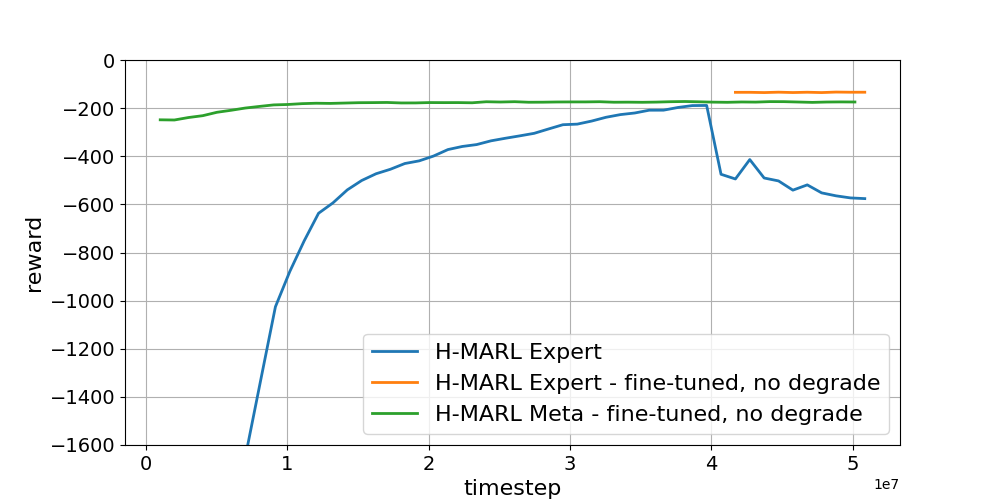}
    \caption{H-MARL with 3 sub-policies: Investigate, Recover, and Control Traffic, after removing the penalty on failed user access (no degrade). All sub-policies are fine-tuned in this new context, against a new red that attempts frequent remote exploits into other subnets. The H-MARL Expert has regained a stable high performance and H-MARL Meta has similar performance.}
    \label{fig:nodegrade}
\end{figure}
 
Defining an expert knowledge to guide the training of a Control Traffic sub-policy to near-optimal is particularly difficult,  due to the conflicting outcomes of using Block actions -- stop red agents from moving through the network, but at the expense of preventing user agents  from completing their work. 

For our current experiments, we use the following expert master policy to train the sub-policies (Algorithm 1), and leave further research into other expert rules for future work: 
\emph{If indicators of compromise are present on hosts, call the Recover sub-policy; otherwise, randomly choose Investigate for 75\% of the time, and Control Traffic for the remaining 25\%.} Thus, the Investigate sub-policy is assigned more weight, as we expect it to be useful more frequently. In fact, for best performance, we use (keep fixed) the Investigate and Recover sub-policies trained previously (see paper), since they have been learned with well-defined expert knowledge, and only train the Control Traffic sub-policy, using  the  rule specified above.

Next, we follow Algorithm 2 to train the Master policy using the three pre-trained sub-policies. Figure~\ref{fig:traffic} shows the reward as training progresses during Algorithm 1 (using the expert rule), and Algorithm 2 (train the master). 
We  observe the instability of the expert, which has been trained with a probability-based rule, reinforcing the importance of learning the master policy. 
The H-MARL Meta algorithm is more stable, as the master policy learns how to combine the sub-policies to solve the meta-task, and is not restricted by a fixed, deterministic rule.

\textbf{Turning off Degrade Service.} In our next set of experiments illustrated in Figure~\ref{fig:nodegrade}, we fine-tuned the sub-policies after turning off the reward penalty of green agents being affected by failed service access. We also used the modified red agent introduced in Section~\ref{sec:experiments_communication} of the supplemental material, which performs remote scanning into other subnets more often, and can still collect rewards by impacting the critical OT service. With block actions being now useful at preventing red agents from spreading, without incurring penalties, H-MARL Expert regains a stable, high performance, as expected. H-MARL Meta achieves similar performance with H-MARL Expert in this setting. The blue agents are using $4\times$ more block actions when the failed user access penalty has been removed.

%% file: supplemental/transfer.tex
\section{H-MARL Transferability}
\label{sec:supl_transfer}
\begin{figure}[th]
    \centering
\includegraphics[width=0.7\linewidth]{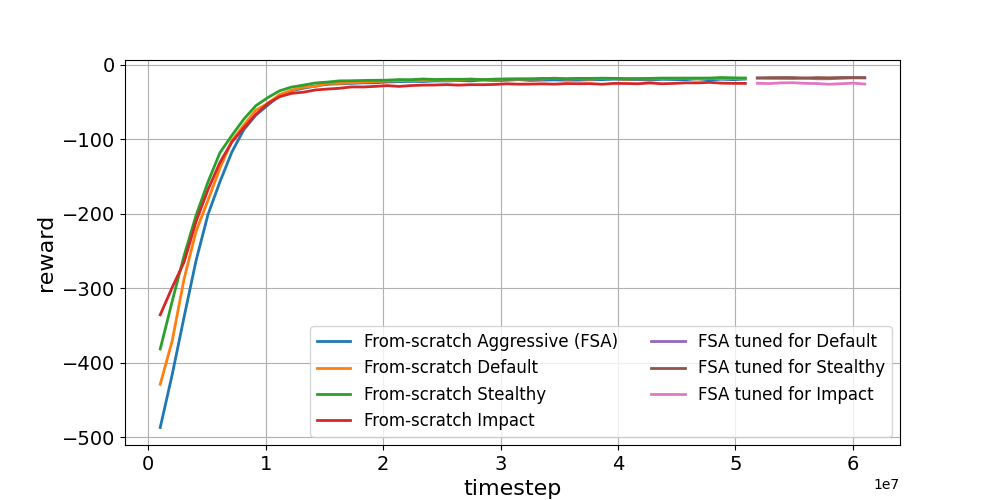}
    \caption{The Recover sub-task is rather agnostic to the attack type and can be re-used against other adversaries.}
    \label{fig:transfer_recover}
\end{figure}

Figure~\ref{fig:transfer_recover} presents  the fine-tuning results of the Recover sub-policy, pre-trained against Aggressive Red, the average-performing attacker. The Recover sub-policy is trained on an observation space consisting of indicators of compromise within a subnet. This sub-policy learns a strong strategy regardless of the attack, and can be directly reused against other red agents.  

